\documentclass[12pt]{article}
\usepackage{amsfonts}
\usepackage{amsmath, amsthm}
\usepackage{epsfig}
\usepackage{rotating}
\usepackage{amsmath,empheq}
\usepackage{authblk}
\usepackage{mathtools}
\usepackage[toc,page]{appendix}

\usepackage{algorithm,algorithmic}
\usepackage{bm}

\newcommand{\dT}{\top}

\usepackage{authblk}

\usepackage{amsmath,verbatim,color,amssymb,epsfig}
\usepackage{bm}
\usepackage{amsfonts}
\usepackage{subfig}
\usepackage{epsfig}
\usepackage{multirow}
\usepackage{graphicx}
\usepackage{array}
\usepackage[table]{xcolor}
\definecolor{Gray}{gray}{0.80}
\usepackage{lscape}
\usepackage{cite}
\usepackage{natbib}
\usepackage[normalem]{ulem}
\usepackage[colorlinks=true,urlcolor=blue,citecolor=purple,linkcolor=blue,bookmarks=true]{hyperref}
\DeclareMathOperator*{\argmax}{argmax} 
\usepackage{caption}
\begin{document}
\def\eqx"#1"{{\label{#1}}}
\def\eqn"#1"{{\ref{#1}}}
\newtheorem{remark}{Remark}
\makeatletter 
\@addtoreset{equation}{section}
\makeatother  

\def\yincomment#1{\vskip 2mm\boxit{\vskip 2mm{\color{red}\bf#1} {\color{blue}\bf --Yin\vskip 2mm}}\vskip 2mm}
\def\squarebox#1{\hbox to #1{\hfill\vbox to #1{\vfill}}}
\def\boxit#1{\vbox{\hrule\hbox{\vrule\kern6pt
          \vbox{\kern6pt#1\kern6pt}\kern6pt\vrule}\hrule}}

\def\theequation{\thesection.\arabic{equation}}
\newcommand{\ds}{\displaystyle}

\newcommand{\bJ}{\mbox{\bf J}}
\newcommand{\bF}{\mbox{\bf F}}
\newcommand{\bM}{\mbox{\bf M}}
\newcommand{\bR}{\mbox{\bf R}}
\newcommand{\bZ}{\mboxZ}
\newcommand{\bX}{\mbox{\bf X}}
\newcommand{\bx}{\mbox{\bf x}}
\newcommand{\bQ}{\mbox{\bf Q}}
\newcommand{\bH}{\mbox{\bf H}}
\newcommand{\bh}{\mbox{\bf h}}
\newcommand{\bz}{\mboxZ}
\newcommand{\ba}{\mbox{\bf a}}
\newcommand{\be}{\mbox{\bf e}}
\newcommand{\bG}{\mboxG}
\newcommand{\bB}{\mbox{\bf B}}
\newcommand{\bb}{\mbox{\bf b}}
\newcommand{\bA}{\mbox{\bf A}}
\newcommand{\bC}{\mbox{\bf C}}
\newcommand{\bI}{\mbox{\bf I}}
\newcommand{\bD}{\mbox{\bf D}}
\newcommand{\bU}{\mbox{\bf U}}
\newcommand{\bc}{\mbox{\bf c}}
\newcommand{\bd}{\mbox{\bf d}}
\newcommand{\bs}{\mbox{\bf s}}
\newcommand{\bS}{\mbox{\bf S}}
\newcommand{\bV}{\mbox{\bf V}}
\newcommand{\bv}{\mbox{\bf v}}
\newcommand{\bW}{\mbox{\bf W}}
\newcommand{\bw}{\mbox{\bf w}}
\newcommand{\bg}{\mboxG}
\newcommand{\bu}{\mbox{\bf u}}
\def\bb{{\bf b}}

\newcommand{\bcU}{\boldsymbol{\cal U}}
\newcommand{\bbeta}{\boldsymbol{\beta}}
\newcommand{\bdelta}{\boldsymbol{\delta}}
\newcommand{\bDelta}{\boldsymbol{\Delta}}
\newcommand{\boldeta}{\boldsymbol{\eta}}
\newcommand{\bxi}{\boldsymbol{\xi}}
\newcommand{\bGamma}{\boldsymbol{\Gamma}}
\newcommand{\bSigma}{\boldsymbol{\Sigma}}
\newcommand{\balpha}{\boldsymbol{\alpha}}
\newcommand{\bOmega}{\boldsymbol{ R}}
\newcommand{\btheta}{\boldsymbol{\theta}}
\newcommand{\bmu}{\boldsymbol{\mu}}
\newcommand{\bnu}{\boldsymbol{\nu}}
\newcommand{\bgamma}{\boldsymbol{\gamma}}

\newtheorem{thm}{Theorem}[section]
\newtheorem{lem}{Lemma}[section]
\newtheorem{rem}{Remark}[section]
\newtheorem{cor}{Corollary}[section]
\newcolumntype{L}[1]{>{\raggedright\let\newline\\\arraybackslash\hspace{0pt}}m{#1}}
\newcolumntype{C}[1]{>{\centering\let\newline\\\arraybackslash\hspace{0pt}}m{#1}}
\newcolumntype{R}[1]{>{\raggedleft\let\newline\\\arraybackslash\hspace{0pt}}m{#1}}

\newcommand{\tabincell}[2]{\begin{tabular}{@{}#1@{}}#2\end{tabular}}
\def\correspondingauthor{\footnote{yixiaoli@hku.hk}}

\title{\bf A Review of Changepoint Detection Models}

\author[1,2]{Yixiao Li\correspondingauthor{}}
\author[1, 3]{Gloria Lin}
\author[1]{Thomas Lau}
\author[1, 4]{Ruochen Zeng}
\affil[1]{Point Zero One Technology}
\affil[2]{University of Hong Kong}
\affil[3]{Imperial College London}
\affil[4]{University of California, Berkeley}
\maketitle

\begin{abstract}
The objective of the change-point detection is to discover the abrupt property changes lying behind the time-series data. In this paper, we firstly summarize the definition and in-depth implication of the changepoint detection. The next stage is to elaborate traditional and some alternative model-based changepoint detection algorithms. Finally, we try to go a bit further in the theory and look into future research directions.
\vspace{0.5cm}
\end{abstract}

\section{Introduction}
Detecting abrupt changes in time-series data has attracted researchers in the statistics and data mining communities for decades \cite{Basseville1993}. Based on the instantaneousness of detection, changepoint detection algorithms can be classified into two categories: online changepoint detection and offline changepoint detection. While the online change detection targets on data that requires instantaneous responses, the offline detection algorithm often triggers delay, which leads to more accurate results. This literature review mainly focuses on the online changepoint detection algorithms.

There are plenty of changepoint detection algorithms that have been proposed and proved pragmatic. The pioneering works \cite{Basseville1993} compared the probability distributions of time-series samples over the past and present intervals. The algorithm demonstrates an abrupt change when two distributions are significantly different. There are various now-famous algorithms following this approach to detect changepoints, such as the generalized likelihood-ratio method \cite{Gustafsson1996} and the change finder \cite{Takeuchi2006}.  Most recently, the subspace methods are proposed, which include subspace identification and Krylov subspace learning \cite{Kawahara2012}.

The aforementioned methods are all considered traditional and rely on pre-designed parametric models, such as the underlying probability distributions, auto-regressive models and state-space models to track specific parameters \cite{Liu2013}. As alternatives, several general and ad-hoc model-free methods have been proposed with no specific parametric assumptions \cite{Desobry2005}. These alternative methods include time-frequency approaches and kernel density estimations. However, a common weakness lies in these algorithms is that they all tend to be less accurate in high-dimensional problems because of the curse of dimensionality \cite{Vapnik1998}. To overcome this problem, we introduce a new strategy called the direct density-ratio estimation.

In summary, this survey focuses on the aforementioned changepoint detection methods and discusses how the algorithms work to detect abrupt changes in details. In Section 2, we explore the traditional model-based changepoint detection algorithms. Section 3 compares the traditional algorithms with the alternative model-free changepoint detections. In Section 4, we make conclusions and present some of the future research directions. 

\section{Model-based Change Detection Algorithms}
\subsection{Generalized Likelihood Ratio}
The generalized likelihood ratio (GLR) test is widely used in detecting abrupt changes in linear systems \cite{Gustafsson1996}, which is proposed by \cite{Basseville1993}. As summarized by \cite{Kerr1987}, the GLR test has an appealing analytical framework that is suitable to those systems with Kalman filters. The test also locates the physical cause of changes when they abruptly occurred. 

In a linear state space model, we present the occurrence of abruptly changes by
$${\bm x}_{t+1}={\bm F}_{t}{\bm x}_t+{\bm G}_t {\bm u}_t+ {\bm w}_t+ \delta(k-t){\bm v},$$
$${\bm y}_t={\bm H}_t {\bm x}_t+{\bm e}_t,$$
where the observation is denoted as ${\bm y}_t$, the input as ${\bm u}_t$, and the state as ${\bm x}_t$. Here, ${\bm w}_t$, ${\bm e}_t$ and ${\bm x}_t$ are assumed to be Gaussian distributed that are mutually independent. The state jump $v$ occurs at an unknown instant $k$. $\delta(j)$ is a pulse function that takes the value of one if $j=0$ and takes the value of zero, otherwise.
The set of measurements ${\bm y}_1, \ldots, {\bm y}_N$ is denoted as ${\bm y}_{1:N}$.

The likelihood function based on the observations up to time $N$ given the jump $\nu$ at time $k$ is denoted ${\bm p(y^N|k,v)}$. The same notation is used for the conditional density function of ${\bm y}^{N}$, where $k$ and $\nu$ are given. The likelihood ratio (LR) test is a multiple hypotheses test, where different jump hypotheses are compared to the no jump null hypothesis in a pairwise manner. In the LR test, the jump magnitude is given. The hypotheses under consideration are
$$H_0: {\rm No\ jump},$$
$$H_1(k,v): {\rm A \ jump\ of\ magnitude\ {\mathnormal v} \ at \ time \ \mathnormal{k}}.$$
By introducing the log-likelihood ratio for the hypotheses test
$$l_N(k,v)\coloneqq2\log\frac{p({\bm y}_{1:N}|H_1(k,v))}{p({\bm y}_{1:N}|H_0)}=2\log\frac{p({\bm y}_{1:N}|k,v)}{p({\bm y}_{1:N}|k=N)},$$
the GLR test is a double optimization over $k$ and $v$
$$\hat{v}(k)=\argmax_v 2\log\frac{p({\bm y}_{1:N}|k,v)}{p({\bm y}_{1:N}|k=N)},$$
$$\hat{k}=\argmax_k 2 \log\frac{p({\bm y}_{1:N}|k,\hat{v}(k))}{p({\bm y}_{1:N}|N)}.$$
The jump candidate $k$ in the GLR test is rejected (a change point is detected), if
$$l_N(\hat{k}, \hat{v}(\hat{k}))>h,$$
where a certain threshold $h$ characterizes the hypothesis test.

\subsection{Bayesian Online Changepoint Detection}
Using the Bayesian approach to detect the abrupt changes in time series has been well studied. In this section, we summarize the works of \cite{Barry1993}, \cite{Paquet2007}, \cite{Adams2007}, and \cite{Garnett2009} to generate a whole picture of the Bayesian approach. 

Let ${\bm y}_1, \ldots, {\bm y}_T$ be a sequence of observations that is divided into non-overlapping product partitions, where the changepoints are the delineations between these partitions. For each partition $\rho$, the data within it are assumed to be $i.i.d.$ generated from a probability distribution $P({\bm y}_t|\eta_\rho )$, while the parameters $\eta_\rho$, $\rho = 1, 2, \ldots$ are assumed to be $i.i.d.$ as well. Define ${\bm y}_t^{(r)}$ as the set of observations associated with the run $r_t$. The Bayesian approach is conducted by estimating the posterior distribution over the current run length  $\{r_t\}$ (i.e., the length of time since the last changepoint), given the data observed
$P({\bm y}_{t+1}|{\bm y}_{1:t})=\sum_{r_t}P({\bm y}_{t+1}|r_t, {\bm y}_t^{(r)})P(r_t|{\bm y}_{1:t}),$
where
\begin{eqnarray*}
P(r_t|{\bm y}_{1:t})&=&\sum_{r_{t-1}}P(r_t,r_{t-1}, {\bm y}_{1:t})\\
&=&\sum_{r_{t-1}}P(r_t,{\bm y}_t|r_{t-1}, {\bm y}_{1:t-1})P(r_{t-1}, {\bm y}_{1:t-1})\\
&=&\sum_{r_{t-1}}P(r_t|r_{t-1})P({\bm y}_t|r_{t-1},{\bm y}_t^{(r)})P(r_{t-1}, {\bm y}_{1:t-1}).\\
\end{eqnarray*}
The model then computes the predictive distribution conditional on $\{r_t\}$ and integrates over the posterior distribution on the current run length to obtain its marginal predictive distribution. A recursive message-passing algorithm is developed for the joint distribution over the current run length and the data, based on two calculations: 1) the prior over $r_t$ given $r_{t-1}$, and 2) the predictive distribution over the newly-observed datum, given the data since the last change point. Furthermore, a recursive algorithm must define not only the recurrence relation but also the initialization conditions. Thus, the prior over the initial run length is the following normalized survival function:
$$P(r_0=\tau)=\frac{1}{Z}S(\tau),$$
$$S(\tau)=\sum_{t=t+1}^\infty P_{gap}(g=t).$$
Furthermore, by addressing the whole problem using the conjugate-exponential models, we have
$${\bm v}_t^{(r)}={\bm v}_{prior}+{\bm r_t},$$
$${\bm \chi}_t^{(r)}={\bm \chi}_{prior}+\sum_{t' \in r_t}{\bm u}(\chi_{t'}).$$
The whole algorithm can be summarized as follows
\begin{algorithm}
\caption{Bayesian Online Changepoint Detection}
\begin{algorithmic}[1]
  
  \STATE Initialize $P(r_0)=\tilde{S}$ or $P(r_0=0)=1$, ${\bm v}_1^{(0)}={\bm v}_{prior}$, ${\bm \chi}_1^{(0)}={\bm \chi}_{prior}$
  \STATE Observe New Datum ${\bm y}_t$
   \STATE Evaluate Predictive Probability $\pi_t^{(r)}=P({\bm y}_t|{\bm v}_t^{(r)}, {\bm x}_t^{(r)})$
   \STATE Calculate Growth Probabilities $P(r_t=r_{t-1}+1, {\bm y}_{1:t})=P(r_{t-1}, {\bm y}_{1:t})\pi_{t}^{(r)}(1-H(r_{t-1}))$
   \STATE Calculate Changepoint Probabilities $P(r_t=0, {\bm y}_{1:t})=\sum_{r_{t-1}}P(r_{t-1}, {\bm y}_{1:t-1})\pi_{t}^{(r)}H(r_{t-1})$
   \STATE Calculate Evidence $P({\bm y}_{1:t})=\sum_{r_t}P(r_t, {\bm y}_{1:t})$
   \STATE Determine Run Length Distribution $P(r_t|{\bm y}_{1:t})=P(r_t, {\bm y}_{1:t})/P({\bm y}_{1:t})$
   \STATE Update Sufficient Statistics ${\bm u}_{t+1}^{(0)}={\bm u}_{prior}$, ${\bm \chi}_{t+1}^{(0)}={\bm \chi}_{prior}$, ${\bm u}_{t+1}^{(r+1)}={\bm u}_t^{(r)}+1$, ${\bm \chi}_{t+1}^{(r+1)}={\bm \chi}_t^{(r)}+{\bm u}(\chi_t)$
   \STATE Perform Prediction $P({\bm y}_{t+1}|{\bm y}_{1:t})=\sum_{r_t}P({\bm y}_{t+1}|{\bm y}_t^{(r)}, r_t)P(r_t|{\bm y}_{1:t})$
   \STATE Return to Step 2
\end{algorithmic}
\end{algorithm}

\subsection{The Subspace Methods for Online Changepoint Detection}
Detecting changepoints in the time-series data based on the subspace identification needs to employ geometric approaches to estimate the linear state-space model \cite{Kawahara2007}. \cite{Takeuchi2006} proposed a framework in which an autoregressive (AR) model is fitted recursively, thereby solving the problems in non-stationary time series. Accordingly, some new changepoint detection algorithms based on the singular-spectrum analysis (SSA) were proposed by \cite{Moskvinz2003}. 

Consider a discrete-time wide-sense stationary vector process ${{\bm y}_{t} \in \bm{R}^p, t = 1, 2, \ldots }$, which models the signal of the unknown stochastic system as a discrete-time linear state-space system:
$${\bm x}_{t+1}={\bm A}{\bm x}_{t}+{\bm v}_{t},$$
$${\bm y}_{t}={\bm C}{\bm x}_{t}+{\bm w}_{t},$$
${\bm x}\in \bm{R}^n$ is a state vector, ${\bm v}\in \bm{R}^n$ and ${\bm w}\in \bm{R}^p$ are the system and observation noises respectively, while ${\bm A}\in \bm{R}^{n\times n}$ and ${\bm C}\in \bm{R}^{p\times n}$ are the system matrices. The key problem solved by the subspace identification is the consistent estimation of the column space of the extended observability matrix.
$$\mathcal{O}_k\coloneqq \left[{\bm C}^\dT({\bm C}{\bm A})^\dT, \ldots, ({\bm C}{\bm A}^{k-1})^\dT\right].$$
Once the extended observability matrix is obtained, we can derive the
system matrices and the Kalman gain by substituting the above equations with 
$${\bm x}_{t+1}={\bm A}{\bm x}_{t}+{\bm K}{\bm e}_{t},$$
$${\bm y}_{t}={\bm C}{\bm x}_{t}+{\bm e}_{t},$$
where ${\bm e}_{t}$ is an innovation process (the error process of the model) and ${\bm K}$ is the stationary
Kalman gain.
Thus, we obtain the extended observability matrix as
$$\mathcal{O}_k=\sum_{ff}^{1/2}{\bm U}_1{\bm S}_1^{1/2}.$$
where the suffix $p$ denotes the past and $f$ denotes the future and the covariance matrices are computed
using the matrices obtained by the $LQ$ factorization, respectively.

A subsequence can be expressed as
$${\bm y}_k(t)=\mathcal{O}_k{\bm x}(t)+\Psi_k{\bm e}_k(t),$$
where ${\Psi_k}$ is defined as 
\[
\Psi_k
\coloneqq
\begin{bmatrix}
    I_{p\times p} & 0 &  \dots  & 0 \\
    {\bm C}{\bm K} & I_{p\times p}   & \dots  & 0 \\
    \vdots & {\bm C}{\bm K}& \vdots & \vdots \\
   {\bm C}{\bm A}^{k-2}K & {\bm C}{\bm A}^{k-3}K & \dots  & I_{p\times p}
\end{bmatrix}.
\]
Moreover, by aligning the above equation according to the structure of a Hankel metrics
$$Y_{k,N}(t)=\mathcal{O}_kX_0+\Psi_kE_{k,N}(t),$$
Hence, the subspace spanned by the column vectors of $Y_{k,N}(t)$ is equivalent to the spans of $\mathcal{O}_k$ plus $\Psi_k$. Then the following distance, which quantifies the gap between
subspaces, can be used as a measure of the changepoint in the time-series
$$\mathcal{D}\coloneqq Y_{k,M}(t_2)^\dT Y_{k,M}(t_2)-Y_{k,M}(t_2)^\dT {\bm U}_1^{(1)}({\bm U}_1^{(1)})^\dT Y_{k,M}(t_2),$$
where ${\bm U}(1)$ is computed by the SVD of the extended observability matrix $\mathcal{O}_k$, which is estimated by the subspace identification using the data in the reference interval
$\mathcal{O}_k^{(1)}={\bm U}^{(1)}{\bm S}^{(1)}({\bm V}^{(1)})^\dT.$
The procedure for change-point detection can be outlined as follows:
\begin{algorithm}
\caption{Subspace Methods for Online Changepoint Detection}
\begin{algorithmic}[1]
\STATE Select $k$, $M$, $N$, $\tau$ and $n$.
\STATE Initialize $P$, $\Sigma_{y1}$, $\Sigma_{y2}$ and $M$.
\STATE At each time $t$
\STATE Update $P$, $\Sigma_{y1}$, $\Sigma_{y2}$ and $M$ by prescribed equations and estimate the observability subspace
\STATE Construct the Hankel matrix ${Y_{k,N}(t2)}$ of the test interval followed by evaluating the distance 
$\mathcal{D}$.
\end{algorithmic}
\end{algorithm}

\section{Alternative Model Free Change Detection Algorithms}
\subsection{Online Kernel Change Detection Algorithm}

In this section, we refer to the famous works written by \cite{Desobry2005} and \cite{Harchaoui2009} to present a general, model-free framework for the online abrupt change detection method called Kernel change detection algorithm. Similar to other model-free techniques, the detection of abrupt changes is based on the descriptors extracted from the signal of interests. 

Let ${\bm y}_1, ..., {\bm y}_n$ be a time series of independent random variables. The change point detection based on the observed sample ${{\bm y}_1, \ldots, {\bm y}_n}$ consists two steps
\begin{itemize}
\item[1)] Decide between
$H_0$: $P({\bm y}_1) = \cdots= P({\bm y}_k) = \cdots = P({\bm y}_n)$ and 
$H_1$: there exists $1 < k^* < n$ such that
$P({\bm y}_1) = \cdots = P({\bm y}_{k^*}),$  $P({\bm y}_{k^*+1}) = \cdots = P({\bm y}_n).$
\item[2)] Estimate $k^*$ from the sample $\{{\bm y}_1, \ldots, {\bm y}_n\}$ if $H_1$ is true.
\end{itemize}

To conduct the kernel changepoint analysis, the running-maximum-partition strategy is employed based on a reproduced  kernel Hilbert space. Let $(\chi, d)$ be a separable measurable metric space, and ${\bm y}$ be a $\chi$-valued random variable with probability measure $\bm{P}$. The expectation with respect to $\bm{P}$ is denoted by $\bm{E}[\cdot]$ while the covariance matrix is denoted by $Cov(\cdot,\cdot)$. Consider a reproducing kernel Hilbert space (RKHS) of function ${\bm \chi}\rightarrow\bm{R}$, the model makes the following two assumptions on the kernel: 1) the kernel $\tau$ is bounded, i.e. $\sup_{(x,y)\in \chi\times\chi} \tau(x,y) < \infty$, 2) for all probability distributions $\bm{P}$, the RKHS associated with $\tau(\cdot,\cdot)$ is dense in $L^2(\bm{P})$.

An efficient strategy for conducting the changepoint analysis is to select the partition of sample. The partition yields a maximum heterogeneity between a sample ${{\bm y}_1, \ldots, {\bm y}_n}$ and a candidate change point $k$ with interval $(1,n)$. Assume that we can compute a measure of heterogeneity ${\Delta_{n,k}}$ between the segments ${{\bm y}_1, \ldots, {\bm y}_k}$ as well as the ${{\bm y}_{k+1}, \ldots, {\bm y}_n}$, then the €œrunning-maximum-partition strategy€ consists in using max ${\Delta_{n,k}}$ as a building block for changepoint analysis. 

Consider a sequence of independent observations ${\bm y_1}, \ldots, {\bm y_n}\in\chi$. For any $[i,j]\subset \{2, \ldots, n-1\}$, the corresponding empirical mean elements and
covariance operators as follows
$$\hat{\bm u}_{i:j}\coloneqq\frac{1}{j-i+1}\sum_{l=i}^j \tau({\bm y}_l, \cdot), \hat{\bm\Sigma}_{i:j}\coloneqq \frac{1}{j-i+1}\sum_{l=i}^j\{\tau({\bm y}_l,\cdot)-\hat{\bm \mu}_{i:j}\}.$$
For all $k\in \{2, \ldots, n -1\}$ the maximum kernel Fisher discriminant ratio (KFDR), is defined as
$$KFDR_{n,k;\gamma}\left({\bm y}_1, \ldots, {\bm y}_n\right)\coloneqq\frac{k(n-k)}{n}\bigg\| \left(\frac{k}{n}\hat{\bm\Sigma}_{1:k}+\frac{n-k}{n}\hat{\bm\Sigma}_{k+1:n}+\gamma I \right)^{-1/2}(\hat{\bm\mu}_{k+1:n})-\hat{\bm\mu}_{1:k}\bigg\|_{\mathcal{H}}^2.$$
This model applies the running-maximum-partition strategy to obtain the building block of the test statistic for change-point analysis. Define the kernel test statistic
$$T_{n;\gamma}(k)\coloneqq\max_{a_n<k<b_n}\frac{KFDR_{n,k;\gamma}-d_{1,n,k;\gamma}(\hat{\bm\Sigma}_{n,k}^W)}{\sqrt{2}d_{2, n, k, \gamma}(\hat{\bm\Sigma}_{n,k}^W)},$$
where $n\hat{\bm\Sigma}_{n,k}^{W}\coloneqq k\hat{\bm\Sigma}_{1:k}+(n-k)\hat{\bm\Sigma}_{k+1:n}.$ The quantities $d_{1, n, k;\gamma}(\hat{\bm\Sigma}_{n,k}^W)$ and $d_{2, n, k;\gamma}(\hat{\bm\Sigma}_{n,k}^W)$, where
$$d_{1, n, k;\gamma}(\hat{\bm\Sigma}_{n,k}^W)=Tr\{(\hat{\bm\Sigma}_{n,k}^W+\gamma I)^{-1}\hat{\bm\Sigma}_{n,k}^W\},$$
$$d_{2, n, k;\gamma}(\hat{\bm\Sigma}_{n,k}^W)=Tr\{(\hat{\bm\Sigma}_{n,k}^W+\gamma I)^{-2}(\hat{\bm\Sigma}_{n,k}^W)^2\},$$
are the normalizing constants for ${T_{n;\gamma}(k)}$ to have zero-mean and unit-variance as $n$ tends to infinity. The maximum is searched within the interval $[a_n, b_n]$ with $a_n > 1$ and $b_n < n$. The algorithm then yields the result of whether an abrupt change has occurred and where the change has occurred.

\subsection{Changepoint Detection by Direct Density Ratio Estimation}
The aforementioned model-free changepoint detection algorithms
tend to be less accurate in high-dimensional problems because of the curse of
dimensionality \cite{Vapnik1998}. To solve the problem, we introduce a new strategy called the direct density-ratio estimation, which estimates the ratio of probability densities directly without
going through density estimation \cite{Liu2013}. Following this idea, models such as the Kullback-Leibler importance estimation procedure (KLIEP) were established \cite{Kawahara2012}. 

Let $\bm{y}(t)$ be a dimensional time series sample at time $t$. The goal of this model is to detect whether there exists a changepoint between two consecutive time intervals, which is
called the reference and test intervals. Let $\bm{Y}_t$ be the forward subsequence of length $k$ at
time $t$
$$\bm{Y}_t=[\bm{y}_t^\dT, \bm{y}_{t+1}^\dT, \ldots, \bm{y}_{t+k-1}^\dT]^\dT.$$
Thus, the likelihood ratio of the sequence sample $\bm{Y}$ is
$$s(\bm{Y})=\log \frac{p_{te}(\bm{Y})}{p_{rf}(\bm{Y})},$$
where ${p_{te}(\bm{Y})}$ and ${p_{rf}(\bm{Y})}$ are the probability density functions of the reference and test sequence samples, respectively.
Let ${\bm t_{rf}}$ and ${\bm t_{te}}$ be the starting points of the reference and test intervals, respectively. Suppose we have ${\bm n_{rf}}$ and ${\bm t_{te}}$ sequence samples in the reference and test intervals. Hence we obtain ${\bm t_{te}}$ = ${\bm t_{rf}}$ + ${\bm n_{rf}}.$ and accordingly, the hypothesis test for this model is given as: 
$$H_0: p(\bm{Y}_i)=p_{rf}(\bm{Y}_i), t_{rf}\leq i<t_{te},$$
$$H_1: p(\bm{Y}_i)=p_{rf}(\bm{Y}_i), t_{rf}\leq i<t; p(\bm{Y}_i)=p_{te}(\bm{Y}_i), t_{te}\leq i<t.$$
The likelihood ratio between the hypotheses $H_0$ and $H_1$ is
\begin{eqnarray*}
\Lambda = \frac{\Pi_{i=1}^{n_{rf}} p_{te}(\bm{Y}_{te}(i))}{\Pi_{i=1}^{n_{rf}} p_{rf}(\bm{Y}_{te}(i)}.
\end{eqnarray*}
Therefore, the model could decide whether there exits a change point between the reference and
test intervals by monitoring the logarithm of the likelihood ratio
$$S(\bm{Y})=\sum_{i=1}^{n_{te}}\log\frac{p_{te}(\bm{Y}_{te}(i))}{p_{rf}(\bm{Y}_{te}(i))}.$$
Based on the logarithm of the likelihood ratio $s(Y)$, the model could detect change occurs if $S\leq{\bm\mu}$.
Thus, we can obtain the density ratio as
$$W(\bm{Y})=\log\frac{p_{te}(\bm{Y})}{p_{rf}(\bm{Y})}.$$
The model solves this problem by using the KLIEP.
The KLIEP first models the density ratio $w(\bm{Y})$ by using a non-parametric Gaussian kernel model
$$\hat{w}(\bm{Y})=\sum_{l=1}^{n_{te}}\alpha_1K_\sigma(\bm{Y}, \bm{Y}_{te}(l)),$$
where $\{\alpha_l\}$ are the parameters to be fitted from samples, and $K_\sigma(\bm{Y}, \bm{Y}')$ is the Gaussian
kernel function with mean $Y'$ and standard deviation $\sigma$
$$K_\sigma(\bm{Y}, \bm{Y}')=\exp\left(-\frac{\|\bm{Y}-\bm{Y}'\|^2}{2\sigma^2}\right).$$
The parameters $\{\alpha_l\}$ in this model are determined such that the empirical Kullback-Leibler divergence from ${p_{te}(\bm{Y})}$ to $\hat{p}_{te}(\bm{Y})$ (= ${p_{rf}(\bm{Y})}$$\hat{w}(\bm{Y})$) is minimized. 

The solution to this problem can be obtained by solving the following convex optimization
problem
\begin{empheq}[left=\empheqlbrace]{align*}
&\max_{\alpha_l}\sum_{i=1}^{n_{te}}\log\left(\sum_{l=1}^{n_{te}}\alpha_l K_\sigma(\bm{Y}_{te}(i), \bm{Y}_{te}(l))\right),\\
& s.t. \ \frac{1}{n_{rf}}\sum_{i=1}^{n_{rf}} \sum_{l=1}^{n_{te}} \alpha_l K_{\sigma}(\bm{Y}_{rf}(i), \bm{Y}_{te}(l))=1,\\
& \alpha_1, \ldots, \alpha_{{n}_{te}}\geq 1.
\end{empheq}
The equality constraint in the above optimization problem comes from the requirement that $\hat{w}(\bm{Y})$ should be properly normalized as $\hat{p}_{te}(\bm{Y})$ (= ${p_{rf}(\bm{Y})}$$\hat{w}(\bm{Y})$) ,which is a probability density function. The non-negativity constraint reflects the non-negativity of the density ratio function.
After solving this optimization problem by arcane procedures, one can detect the change
points in a data series by the following algorithm:
\begin{algorithm}
\caption{Changepoint Detection by Direct Density Ratio Estimation}
\begin{algorithmic}[1]
  \STATE Input: New sample $\bm{y}(t)$, the previous estimate of parameters $\bm{\alpha}$ and forgetting factors $\eta$ and $\lambda$.
  \STATE Create new sequence of sample $\bm{Y}_{te}(n_{te}(n_{te}+1)).$
   \STATE Update the parameters $\bm{\alpha}$:  
   \[
\bm{\alpha} \leftarrow
\begin{bmatrix}
    (1-\eta\lambda)\alpha_2 &   \\
     (1-\eta\lambda)\alpha_3&      \\
    \vdots &   \\
   (1-\eta\lambda)\alpha_{n_{te}} &  \\
   \eta/c
\end{bmatrix},
\] \\
where $c=\sum_{l=1}^{n_{te}}\alpha_l K_\sigma(\bm{Y}_{te}(n_{te}+1), \bm{Y}_{te}(l))$.
\STATE Perform feasibility satisfaction:
\begin{eqnarray*}
&&\bm{\alpha}\leftarrow\bm{\alpha}+(1-\bm{b}^\dT\bm{\alpha})\bm{b}/(\bm{b}^\dT\bm{b}),\\
&&\bm{\alpha}\leftarrow\max(\bm{0},\bm{\alpha}),\\
&&\bm{\alpha}\leftarrow\bm{\alpha}(\bm{b}^\dT\bm{\alpha}),
\end{eqnarray*}
where $b_l=\frac{1}{n_{rf}}\sum_{i=1}^{n_{rf}}K_\sigma(\bm{Y}_{rf}(i), \bm{Y}_{te}(l))$ for $l=1,\ldots, n_{rf}.$
\STATE Update as $\bm{Y}_{rf}(n_{rf}+1)\leftarrow \bm{Y}_{te}(1).$
\end{algorithmic}
\end{algorithm}

\section{Conclusion}
Changepoint detection has always been a subject worth of studying and exploring. There is a flourish of old literature and traditional models devoted to this subject. Throughout these years, more and more new methodologies have been introduced to tackle the abrupt changes in data series. In this literature review, we have summarized a portion of the most famous and effective methods to detect change point. As for future research directions, the academia is now heading to find more methods based on non-parametric model-free algorithms to detect change points, such as the single spectrum method, direct density estimation method, etc.

\end{document}